\documentclass[dvipsnames]{article}

\usepackage[final]{lychee}
\usepackage[utf8]{inputenc} 
\usepackage[T1]{fontenc}    
\usepackage{url}            
\usepackage{booktabs}       
\usepackage{amsfonts}       
\usepackage{nicefrac}       
\usepackage{multirow}
\usepackage{wrapfig}
\usepackage{amssymb}
\usepackage{epigraph}
\usepackage{makecell}
\usepackage{listings}
\usepackage{subcaption}     
\usepackage[labelfont=bf]{caption} 
\usepackage[misc]{ifsym}
\usepackage[ruled,vlined]{algorithm2e}
\usepackage{color, soul}
\usepackage{microtype}      
\usepackage{xcolor}         
\usepackage{geometry}
\usepackage{times}
\usepackage{ragged2e}
\usepackage{amsmath}
\usepackage{bm}
\usepackage{latexsym}
\usepackage{graphicx}
\usepackage{float}
\usepackage{longtable}
\usepackage{tabu}
\usepackage{bbding}
\usepackage{awesomebox}
\usepackage[most,breakable]{tcolorbox}
\usepackage{etoolbox}
\usepackage{fancyhdr}
\usepackage{lipsum}  
\usepackage{CJKutf8}
\usepackage{pdfpages}
\usepackage{multicol}
\usepackage{pgffor} 
\usepackage{fontawesome5}
\usepackage{nicematrix} 
\usepackage{paralist}
\usepackage[edges]{forest}
\usepackage{siunitx}
\usepackage{tikz-dependency}
\usepackage{tikz}
\usepackage{bbm}
\usepackage[english]{babel}  
\usepackage{hyphenat}

\usepackage{inconsolata}     
\usepackage{array}           
\usepackage{arydshln}        
\usepackage{enumitem}        
\newtcolorbox{promptbox}[1]{
    colback=gray!3!white,       
    colframe=black!75,         
    coltitle=white,           
    title={\sffamily\bfseries #1}, 
    fontupper=\small,           
    boxrule=0.8pt,             
    arc=3pt,                   
    left=8pt, right=8pt, top=6pt, bottom=6pt, 
    before skip=10pt,          
    after skip=10pt            
}

\definecolor{promptgray}{HTML}{EFEFEF}

\newcommand{\Rmnum}[1]{\expandafter\@slowromancap\romannumeral #1@}

\usepackage{caption} 
\usepackage[colorlinks, linkcolor=RoyalBlue, anchorcolor=BrickRed, citecolor=RoyalBlue, urlcolor=RoyalBlue]{hyperref}
\usepackage{cleveref}
\crefname{section}{§}{§§}
\Crefname{section}{§}{§§}

\definecolor{msftBlue}{RGB}{0,164,239}
\definecolor{msftGreen}{RGB}{127,186,0}
\definecolor{msftYello}{RGB}{255,185,0}
\definecolor{mypurple}{RGB}{138,43,226} 
\definecolor{msftBlack}{RGB}{0,0,0}

\newtcolorbox{myboxnote}[1][]{
  breakable,
  title=#1,
  colback=cyan!0,
  colbacktitle=cyan!0,
  coltitle=black,
  fonttitle=\bfseries,
  bottomrule=0pt,
  toprule=0pt,
  leftrule=1.5pt,
  rightrule=1.5pt,
  titlerule=0pt,
  arc=0pt,
  outer arc=0pt,
  colframe=lightgray,
}
\usepackage{mdframed}
\definecolor{academicblue}{RGB}{54, 95, 145}

\tcbset{
  iclrtakeawaybox/.style={
    width=\linewidth,
    colback=gray!5,
    colframe=gray!60,
    colbacktitle=gray!30,
    coltitle=black,
    fonttitle=\bfseries,
    boxrule=0.5pt,
    arc=2pt,
    sharp corners=south,
    attach boxed title to top left={yshift=-0.1in,xshift=0.15in},
    boxed title style={boxrule=0pt, colframe=white},
    enhanced,
    drop shadow southeast
  }
}
\newtcolorbox{TakeawayBox}[2][]{iclrtakeawaybox,title=#2,#1}

\newenvironment{itemsize*}%
 {\leftmargini=20pt\begin{itemize}%
  \setlength{\itemsep}{3pt}%
  \setlength{\parskip}{0pt}%
  }%
 {\end{itemize}}
\newenvironment{enumerate*}%
 {\begin{enumerate}%
  \setlength{\itemsep}{0pt}%
  \setlength{\parskip}{0pt}}%
 {\end{enumerate}}

\usepackage{xparse}
\usepackage{colortbl}

\usepackage{svg}

\title{
\fontsize{16}{19}\selectfont Hierarchical Acoustic-Semantic Modeling:  Modality \\ Separation and Semantic Coherence for Full-Duplex SLMs}

\author{
 \textbf{Zhenyu Liu\textsuperscript{1,2}},
 \textbf{Xuanyu Zhang\textsuperscript{1,2}},
 \textbf{Yunxin Li\textsuperscript{1,2}},
 \textbf{Qixun Teng\textsuperscript{1}},
 \textbf{Shenyuan Jiang\textsuperscript{1}},
 \textbf{Haolan Chen\textsuperscript{}},\\
 \textbf{Minjun Zhao\textsuperscript{}},
 \textbf{Fanbo Meng\textsuperscript{}},
 \textbf{Yu Xu\textsuperscript{}},
 \textbf{Yancheng He\textsuperscript{}},
 \textbf{Baotian Hu\textsuperscript{1,2,~\Letter}\thanks{\Letter\ Corresponding author.}}, 
 \textbf{Haizhou Li\textsuperscript{2,3}},
 \textbf{Min Zhang\textsuperscript{1,2}}
\\~
\textsuperscript{1}School of Computer Science and Technology, Harbin Institute of Technology, Shenzhen, China\\
\textsuperscript{2}Center for Language, Intelligence and Machines, Shenzhen Loop Area Institute, Shenzhen, China\\
\textsuperscript{3}School of Artificial Intelligence, The Chinese University of Hong Kong, Shenzhen, China \\
\\~
\\
\href{https://huggingface.co/HIT-TMG/Lychee-FD}{\includegraphics[height=1em]{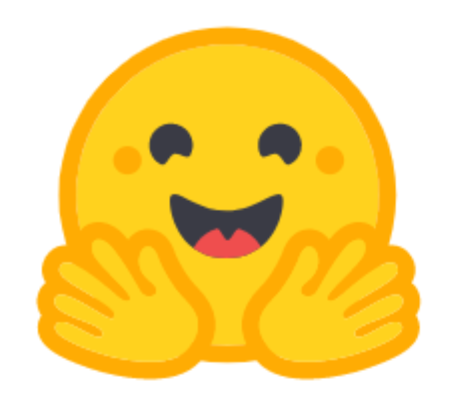} https://huggingface.co/HIT-TMG/Lychee-FD} \\
\href{https://github.com/HITsz-TMG/Lychee-FD}{\includegraphics[height=1em]{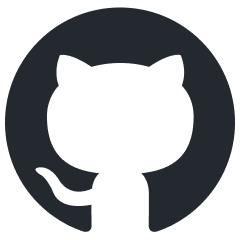} https://github.com/HITsz-TMG/Lychee-FD} \\
\href{https://hitsz-tmg.github.io/Lychee-FD}{\includegraphics[height=1em]{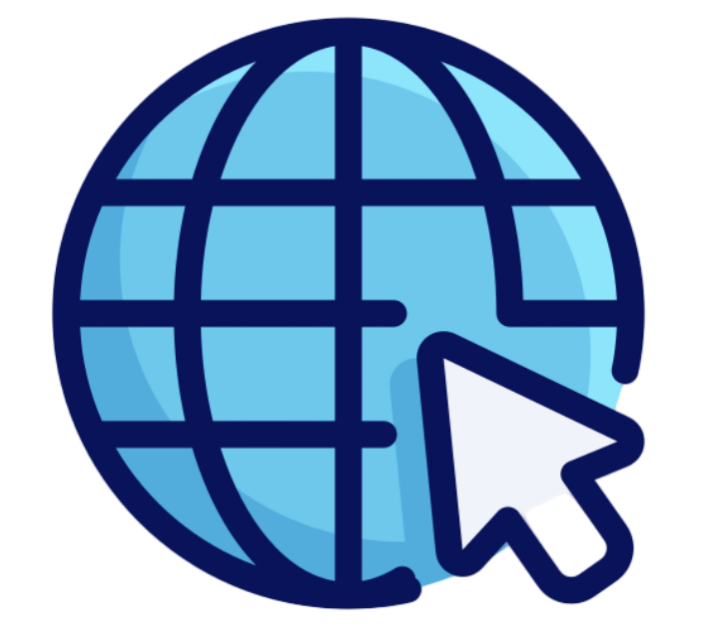} https://hitsz-tmg.github.io/Lychee-FD}
\\~\\
\small{
  \textbf{Correspondence:} 
  \href{mailto:liuzhenyuhit@gmail.com}{liuzhenyuhit@gmail.com},
  \href{mailto:25s151197@stu.hit.edu.cn}{25s151197@stu.hit.edu.cn},
  \href{mailto:liyx@hit.edu.cn}{liyx@hit.edu.cn}
}
\\
\small{
  \href{mailto:hubaotian@hit.edu.cn}{hubaotian@hit.edu.cn},
  \href{mailto:zhangmin2021@hit.edu.cn}{zhangmin2021@hit.edu.cn},
  \href{mailto:haizhouli@cuhk.edu.cn}{haizhouli@cuhk.edu.cn}
}
}

\begin{document}

\maketitle
\begin{abstract}
Developing seamless, high-performance, native intelligent full-duplex Spoken Language Models (SLMs) remains a critical challenge and long-standing goal for the speech and NLP community. 
Despite notable progress, recent endeavors are fundamentally constrained by severe \textbf{modality interference}, which causes substantial knowledge degradation and compromises semantic integrity --- ultimately making full-duplex SLMs feel unnatural and unintelligent.
In this paper, through an exhaustive fine-grained analysis of model optimization dynamics, we uncover the root cause of such performance degradation, revealing that modality interference arises from \textbf{inherent gradient conflicts} between acoustic and semantic modeling when the two modalities are forced to share a deep parameter space. 
Guided by this key insight, we introduce \textbf{Lychee-FD}, a native end-to-end full-duplex framework designed to mitigate modality interference. Importantly, we propose a hierarchical parameter separation strategy that decouples conflicting modalities in deep layers while preserving cross-modality coherence via a dedicated semantic alignment channel. 
Extensive experiments on multiple full-duplex benchmarks demonstrate that our method significantly advances the state of the art, yielding substantial improvements in both speech intelligence (+\textbf{7.4\%} on Spoken QA) and full-duplex interaction fluidity (+\textbf{28.5\%} on FullDuplexBench 1.5) without compromising inference efficiency. 
To the best of our knowledge, this work is the first to achieve two key advances: 1) uncovering and elucidating the root cause of modality interference in full-duplex SLMs, and 2) designing an elegant hierarchical model together with a practical solution for seamless, high-performance, native intelligent full-duplex SLMs.
\end{abstract}

\section{Introduction}

The rapid evolution of Large Language Models (LLMs) has fundamentally reshaped our daily lives, establishing them as ubiquitous assistants capable of complex reasoning and instruction following. Within this landscape, Spoken Language Models (SLMs) represent a significant paradigm shift from text-based to voice-based interaction. Despite recent advancements in Omni-modal models capable of seamless voice interaction~\citep{openai2024gpt4ocard,DBLP:conf/acl/ZhanDYZZLZYZL0F24,li2025uni,DBLP:journals/corr/abs-2001-08361,10.5555/3600270.3602446,DBLP:journals/pami/XuZC23,DBLP:journals/corr/abs-2501-16327}, a critical disparity remains between artificial agents and authentic human conversation. Currently, most voice interactions are constrained to a rigid half-duplex mode, where the system strictly alternates between listening and speaking in a sequential manner. In contrast, authentic human conversation is inherently full-duplex, requiring the ability to continuously process incoming audio streams while concurrently generating responses~\citep{73837945-26a5-3e38-a7b9-daf138681621,DBLP:journals/corr/abs-2501-01957,DBLP:journals/corr/abs-2411-13577}. The forced turn-taking of half-duplex systems disrupts the fluidity of interaction, creating an artificial barrier between user and agent.


\begin{wrapfigure}{r}{0.48\textwidth}

\centering
\includegraphics[width=0.98\linewidth]{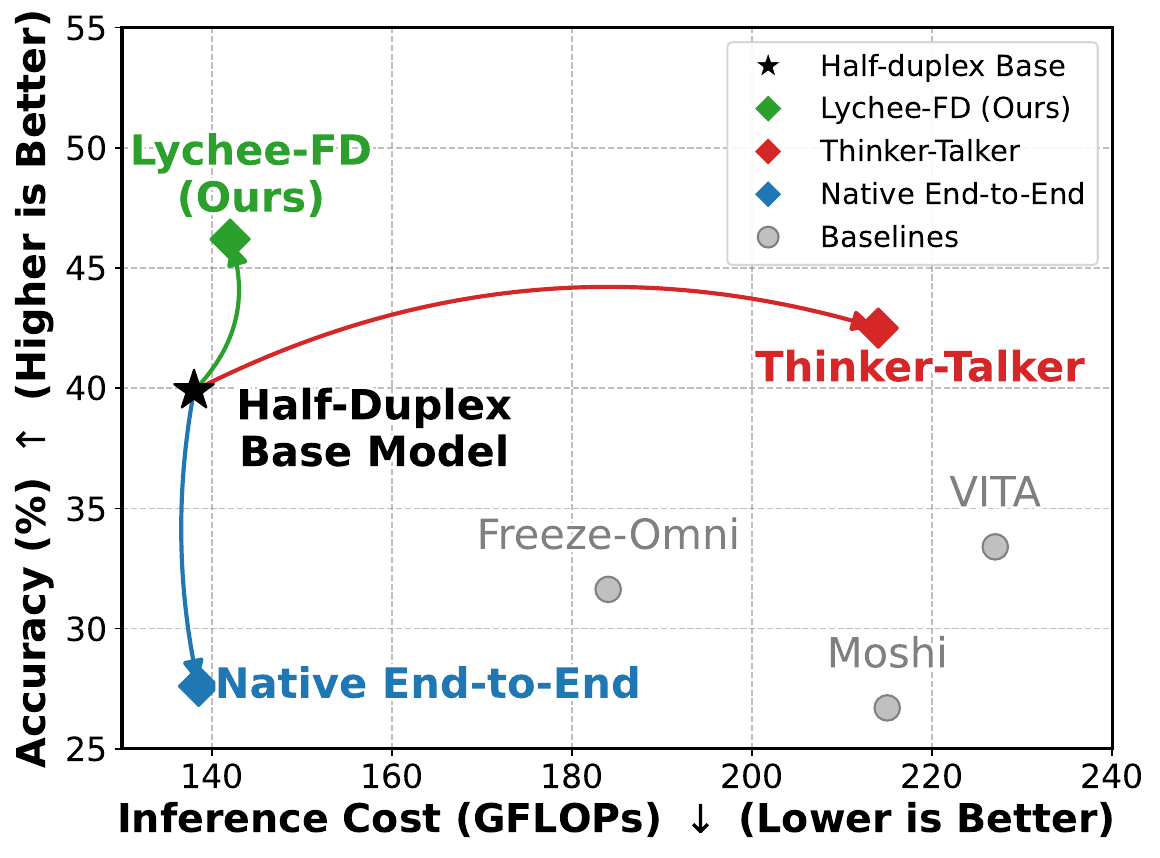}
\caption{Visualization of the efficiency and intelligence trade-off. Existing paradigms face a dilemma when extending a half-duplex SLM (\textbf{\textcolor{black}{black star}}) to a full-duplex one. Native End-to-End models (\textbf{\textcolor{RoyalBlue}{blue diamond}}) sacrifice accuracy for efficiency, while Thinker-Talker models (\textbf{\textcolor{BrickRed}{red diamond}}) preserve knowledge but incur prohibitive inference costs. In contrast, our proposed Hierarchical framework (\textbf{\textcolor{ForestGreen}{green diamond}}) combines low latency with high accuracy, significantly outperforming all full-duplex SLM baselines.}
\label{fig:introduction}
\end{wrapfigure}

To bridge this gap, developing native Full-Duplex SLMs (FDSLMs) has emerged as a critical challenge and a long-standing goal for the speech and NLP communities~\citep{73837945-26a5-3e38-a7b9-daf138681621,DBLP:journals/tmlr/AroraCCPWADLLW25,DBLP:journals/tacl/NguyenKCAHETASM23,DBLP:conf/kdd/LinWHSSL22}. This full-duplex paradigm demands complex pragmatic reasoning, enabling the agent to handle spontaneous interruptions, inject natural backchannels, and dynamically manage turn-taking without explicit system-level triggers~\citep{DBLP:conf/interspeech/HuHCCGZCLBG25, DBLP:conf/chi/LiuSXHYZ025}. Consequently, achieving seamless full-duplex communication is widely regarded as the next critical milestone in human-machine interaction, promising to unlock a truly natural, fluid, and immersive conversation experience~\citep{DBLP:journals/corr/abs-2502-09940,DBLP:journals/corr/abs-2501-01957,DBLP:journals/corr/abs-2503-04721}.

Despite progress in FDSLM development~\citep{DBLP:journals/corr/abs-2408-05211,DBLP:conf/iclr/ZengDLZJD025}, current methods still suffer from severe \textbf{Modality Interference}. As illustrated in Figure~\ref{fig:introduction}, adapting a half-duplex base model (black star) into a native End-to-End architecture (blue diamond) precipitates significant knowledge degradation. This degradation is further corroborated by state-of-the-art models such as Moshi~\citep{DBLP:journals/corr/abs-2410-00037}, which reports a significant drop in speech intelligence (-12.7\% on LlamaQ) after full-duplex alignment. Historically, this type of interference has been a notorious and persistent optimization bottleneck across deep learning~\citep{DBLP:journals/corr/Ruder17a,Caruana1997,NEURIPS2020_3fe78a8a,NEURIPS2018_432aca3a}. While some approaches~\citep{DBLP:conf/icml/WangLFZS000M25,chen2025funaudiochattechnicalreport} attempt to circumvent this interference by adopting Thinker-Talker architectures (red diamond), they require intricate multi-stage training and introduce significant latency. These limitations indicate that existing paradigms either fail in knowledge retention or inference efficiency. Consequently, the central research question of this work is: \textit{How can we resolve this modality interference to simultaneously achieve high inference efficiency and robust knowledge retention in full-duplex SLMs?}

To uncover the root cause of modality interference, we conduct an exhaustive fine-grained \textbf{model optimization dynamics analysis} (shown in Figure~\ref{fig:motivation}). By quantifying the geometric relationships between the gradient vectors of semantic and acoustic objectives, we demonstrate the \textbf{inherent gradient conflicts} that fundamentally cause this interference. 
First, by calculating the layer-wise gradient cosine similarity, we reveal severe \textbf{optimization divergence} in gradient directions. While the gradient directions of text and speech objectives are synergistic in shallow layers, they become increasingly orthogonal and negative in deeper layers. This empirically demonstrates that forcing acoustic and semantic modeling to update within a shared parameter space inevitably fractures the optimization trajectory. Second, by evaluating the gradient magnitude ratio, we identify severe \textbf{semantic dilution}. The temporal alignment of sparse text tokens (e.g., 3Hz) with dense audio frames (e.g., 25Hz) via padding tokens consistently suppresses the magnitude of semantic gradients across all layers. Consequently, the optimization landscape becomes overwhelmingly dominated by acoustic modeling, effectively suppressing semantic modeling.

Inspired by these insights, we introduce \textbf{Lychee-FD}, a native end-to-end full-duplex framework designed to mitigate modality interference with two architectural innovations.
First, we propose a \textbf{hierarchical parameter separation} strategy for optimization divergence in deep layers. Specifically, we separate the deep layers into independent acoustic and semantic heads. By executing these heads in parallel, we maintain the original model depth, thereby preserving inference efficiency.
Second, to counter semantic dilution, we introduce a \textbf{semantic alignment channel} to generate coherent internal monologues. By utilizing continuous textual supervision as a semantic anchor, we preserve the robustness of semantic modeling during training. 
Extensive experiments demonstrate that Lychee-FD achieves the state-of-the-art performance, specifically delivering an average 7.4\% improvement on Spoken QA tasks and a 28.5\% gain on FullDuplexBench 1.5. 
Ultimately, our approach effectively mitigates modality interference, simultaneously achieving high inference efficiency and robust knowledge retention.

Our contributions are summarized as follows:
\begin{itemize}
    \item For the first time, we uncover the root cause of modality interference in full-duplex SLMs. Supported by an exhaustive fine-grained analysis of model optimization dynamics, we reveal that these obstacles stem from \textbf{inherent gradient conflicts} between acoustic and semantic modeling when forced into a shared deep parameter space.
    
     \item We present \textbf{Lychee-FD}, the first fully native intelligent full-duplex framework. We propose a hierarchical parameter separation strategy that decouples acoustic and semantic modeling, bridged by an elegantly designed semantic alignment channel to preserve coherent internal monologues and robust semantics.
    
    \item We advance the state-of-the-art across multiple full-duplex benchmarks. To forge ahead the full-duplex research in the communities, we open-source our framework, training pipeline, and model weights at \url{https://github.com/HITsz-TMG/Lychee-FD}.
    
\end{itemize}

\section{Related Work}

\subsection{Spoken Language Model}

SLMs have evolved from cascading ASR-LLM-TTS pipelines~\citep{DBLP:conf/emnlp/ZhangLZZWZQ23,DBLP:journals/corr/abs-2407-04051,DBLP:journals/taslp/BorsosMVKPSRTGTZ23,10842513,DBLP:journals/taslp/ZeghidourLOST22} to unified architectures. Current methods are mainly grouped into two paradigms:

\paragraph{Thinker-Talker Architectures.} This paradigm decouples acoustic generation from the LLM backbone to mitigate modality interference~\citep{DBLP:journals/corr/abs-2503-20215,DBLP:conf/icml/WangLFZS000M25,DBLP:conf/iclr/FangGZMZ025}. For instance, \citet{DBLP:journals/corr/abs-2503-20215} adopts a dual-track autoregressive architecture. Despite their stability, these systems often require intricate, multi-stage training curricula and suffer from inference bottlenecks caused by the separate generation modules.

\paragraph{Native End-to-End Architectures.} Conversely, these architectures embed speech and text into a shared semantic space, enabling direct speech-to-speech reasoning~\citep{li2025perception,DBLP:journals/corr/abs-2408-16725,DBLP:conf/emnlp/MitsuiMWHS24,DBLP:conf/iclr/ZengDLZJD025}. For examples, Step-Audio~\citep{DBLP:journals/corr/abs-2507-16632} introduces latent audio encoding to capture paralinguistic cues. From a complementary perspective, MOSS-Speech~\citep{DBLP:journals/corr/abs-2501-16327} proposes a modality-based layer-splitting architecture through a layer-wise analysis of speech--text representations, using modality-specific upper branches for speech and text generation.

Crucially, while existing SLMs remain fundamentally bottlenecked by rigid, half-duplex turn-taking mechanisms, our work advances the field by establishing a native Full-Duplex framework. By enabling simultaneous listening and speaking, we aim to unlock a truly natural, fluid, and immersive paradigm for future human-machine interaction.

\subsection{Full Duplex Speech Interaction}

To achieve turn-free and fluid conversation, recent research has explored full-duplex paradigms that transcend rigid turn-taking. These efforts can be broadly categorized into three paradigms:
\paragraph{Full-Duplex Dialogue System.} Early efforts primarily leverage Voice Activity Detection (VAD) as a dialogue manager to control the speaking process of half-duplex SLMs~\citep{DBLP:journals/corr/abs-2509-06502,DBLP:journals/corr/abs-2502-13472,DBLP:journals/corr/abs-2509-23938,DBLP:journals/corr/abs-2502-14145}. Although making some progress, these cascaded systems suffer from high latency and error propagation, prompting a shift towards unified modeling.

\paragraph{Time-Division Multiplexing (TDM).} To address these limitations, recent works use the SLM itself as the dialogue manager. TDM approaches flatten listening and speaking tokens into a single temporal sequence. However, this leads to increasing computational complexity and limits long-context interaction~\citep{DBLP:conf/acl/ZhangCDCWZLYTDZ25,DBLP:conf/emnlp/VeluriPYGG24,DBLP:conf/emnlp/ZhangCH0XXZ0024,DBLP:journals/corr/abs-2505-17060}. 

\paragraph{Channel-Division Multiplexing (CDM).} Instead of flattening, CDM approaches explicitly model concurrent input and output streams, offering the most integrated form of interaction~\citep{coreteam2025mimoaudioaudiolanguagemodels,DBLP:journals/tacl/NguyenKCAHETASM23,DBLP:journals/corr/abs-2410-11190}. For example, \citet{DBLP:journals/corr/abs-2410-00037} and \citet{chen2025funaudiochattechnicalreport} integrate time-aligned text and audio streams to provide explicit semantic guidance for speech generation. However, these fully shared architectures entangle distinct modalities in a single parameter space, leaving them highly susceptible to severe knowledge degradation.

Unlike previous works that struggle with modality interference, Lychee-FD provides the first fundamental solution to the core problem within FDSLMs. By resolving the inherent deep-layer gradient conflicts between acoustic and semantic modeling, our native end-to-end framework successfully achieves ultra-low latency with robust knowledge retention, substantially pushing forward the frontier of research for the speech and NLP communities.


\begin{figure}[ht!]
    \centering
    \begin{subfigure}[b]{0.48\textwidth}
        \centering
        $\vcenter{\hbox{\includegraphics[width=\linewidth]{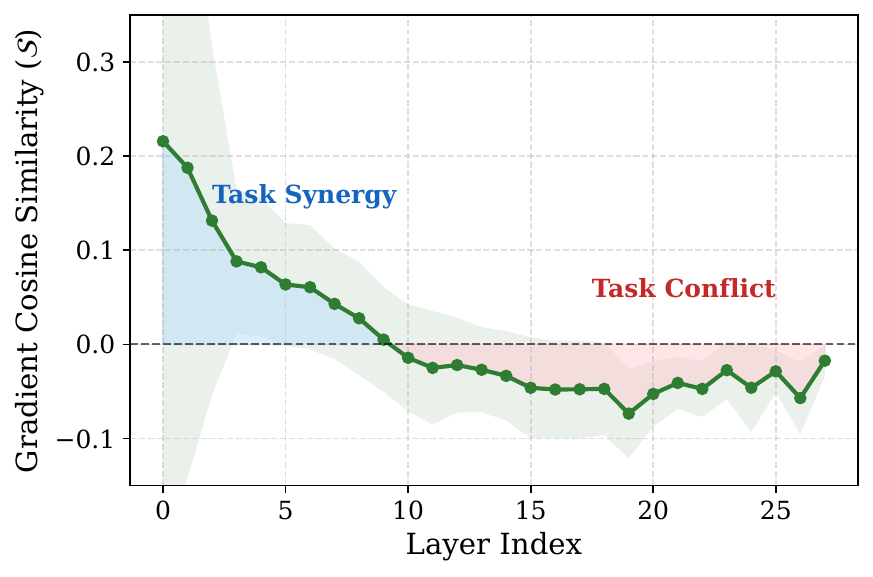}}}$
        \caption{Gradient Cosine Similarity}
        \label{fig:motivation_sim}
    \end{subfigure}
    \hfill 
    \begin{subfigure}[b]{0.48\textwidth}
        \centering
        $\vcenter{\hbox{\includegraphics[width=\linewidth]{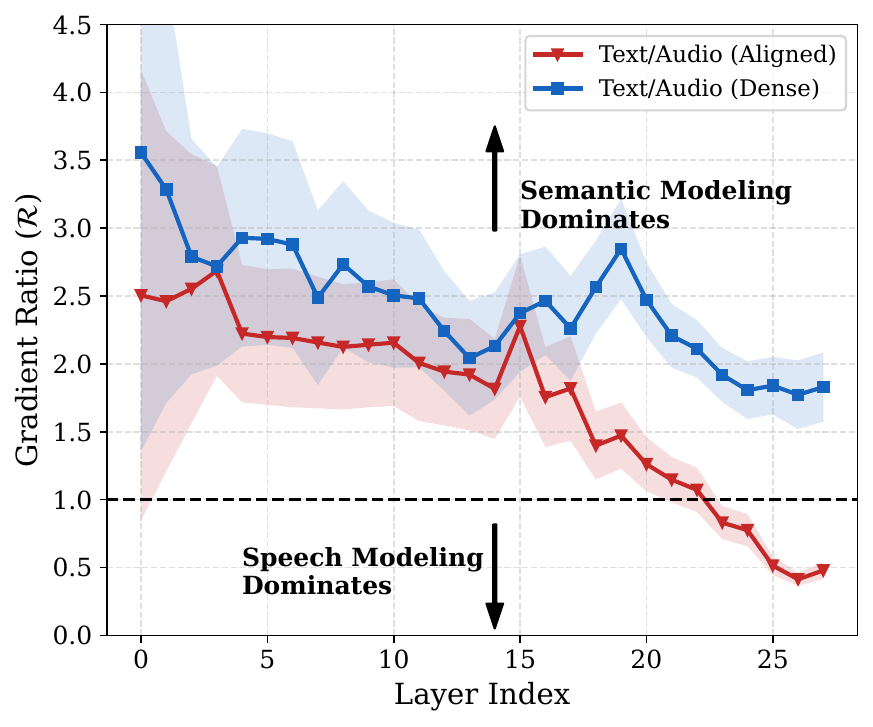}}}$
        \caption{Gradient Magnitude Ratio}
        \label{fig:motivation_radio}
    \end{subfigure}
    
    \caption{\textbf{Optimization Dynamics Visualization.} \textbf{(a) Gradient Cosine Similarity:} The transition to negative values in deep layers reveals conflicting optimization directions between semantic and acoustic modeling, motivating our hierarchical parameter separation. \textbf{(b) Gradient Magnitude Ratio:} The consistently lower ratio for ``Aligned'' (Red) compared to ``Dense'' (Blue) indicates that sparse alignment dilutes semantic supervision, motivating our semantic alignment channel.}
    \label{fig:motivation}
\end{figure}

\section{Hierarchical Acoustic-Semantic Modeling}

\subsection{Model Optimization Dynamics Analysis of Modality Interference}
\label{sec:motivation}

To empirically investigate the fundamental root cause of the modality interference and understand the learning process within FDSLMs, we conducted an in-depth model optimization dynamics analysis, shown in Figure~\ref{fig:motivation}. We utilized a native CDM architecture initialized with weights from StepAudio-2-mini~\citep{DBLP:journals/corr/abs-2507-16632}. We performed forward passes on 1K samples from the training set to accumulate gradients for the cross-entropy losses of both text token generation  ($\mathcal{L}_{\text{text}}$) and speech token generation ($\mathcal{L}_{\text{speech}}$), without updating the model parameters. Specifically, the layer-wise gradient vectors are formally defined as follows:

\begin{equation}
    \mathbf{g}_{\text{text}}^{(l)} = \nabla_{\theta^{(l)}} \mathcal{L}_{\text{text}},
\end{equation}
\begin{equation}
    \mathbf{g}_{\text{speech}}^{(l)} = \nabla_{\theta^{(l)}} \mathcal{L}_{\text{speech}},
\end{equation}
where $\nabla_{\theta^{(l)}}$ denotes the gradient operator with respect to the layer parameters $\theta^{(l)}$, and the resulting tensors are flattened into vectors. By analyzing the geometric relationships of these gradient vectors, we quantitatively uncover the fundamental interaction dynamics between the two modalities during the entire learning process.

\paragraph{Optimization Divergence.} We investigated the compatibility of the two optimization objectives by calculating the cosine similarity $\mathcal{S}^{(l)}$ between text and speech gradient vectors:
\begin{equation}
    \mathcal{S}^{(l)} = \cos(\mathbf{g}_{\text{text}}^{(l)}, \mathbf{g}_{\text{speech}}^{(l)}). 
\end{equation}
As shown in Figure~\ref{fig:motivation_sim}, the similarity reveals a distinct layer-wise pattern. In the shallow layers (0-9), the cosine similarity is positive, indicating that the two modalities share synergistic optimization directions, focusing on common low-level features processing. However, as the depth increases, the similarity drops sharply, turning negative and fluctuating in the deeper layers. This trend empirically confirms our hypothesis regarding the dual nature of speech: while shallow layers can share representations, the deep layers face a fundamental conflict between acoustic and semantic modeling. Forcing a unified set of parameters to resolve these opposing gradient directions inevitably leads to sub-optimal performance, validating the root cause of the observed modality interference. Moreover, we confirm these findings on Moshi architecture in Section~\ref{app:moshi_dynamics} and provide a global gradient influence analysis in in Appendix~\ref{app:gradient_influence} to demonstrate the causal destructive impact of this modality conflict.

\begin{figure*}[ht!]
\centering
\includegraphics[width=\linewidth]{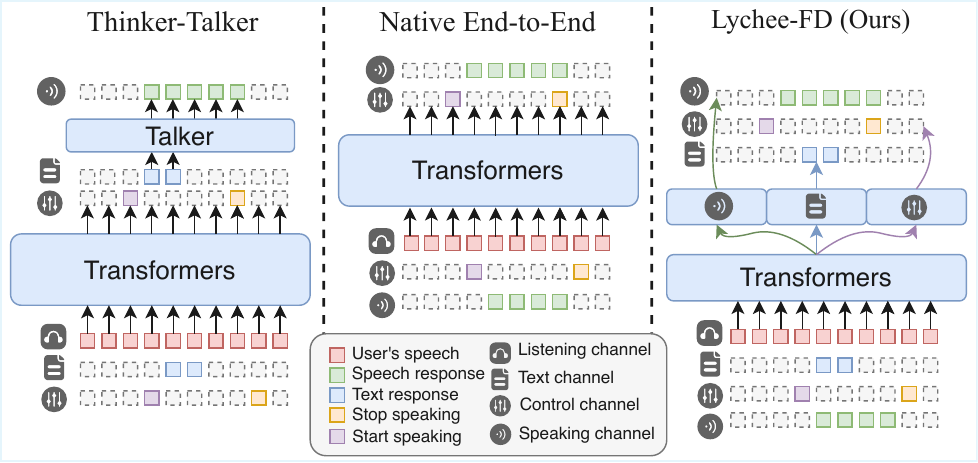}
\caption{Two mainstream architecture paradigms of SLMs and our proposed Lychee-FD. Our design features a hierarchical parameter separation strategy to resolve deep-layer modality conflicts and a semantic alignment channel to enforce robust knowledge retention without sacrificing inference efficiency.}
\label{fig:architecture}
\end{figure*}

\paragraph{Semantic Dilution.} Prevalent Full-Duplex SLMs typically address the frequency mismatch between text (approximately 3Hz) and audio (typically 25Hz) by interleaving padding tokens to enforce temporal alignment~\citep{DBLP:journals/corr/abs-2410-00037,DBLP:journals/corr/abs-2507-16632,chen2025funaudiochattechnicalreport}. To evaluate the impact of this alignment on optimization, we compared the ratio $\mathcal{R}^{(l)}$ of gradient magnitudes between the two modalities: 
\begin{equation}
    \mathcal{R}^{(l)} = ||\mathbf{g}_{\text{text}}^{(l)}|| / ||\mathbf{g}_{\text{speech}}^{(l)}||. 
\end{equation}
As illustrated in Figure~\ref{fig:motivation_radio}, we observe a substantial disparity between the continuous text supervision (Dense) and the sparse text with padding (Aligned). Specifically, the gradient magnitude ratio in the Aligned setting is consistently suppressed across all layers, suggesting that the introduction of padding tokens effectively dilutes the density of semantic supervision. Consequently, the optimization dynamics become dominated by acoustic reconstruction, which drives the observed degradation in knowledge retention.

\subsection{Model Design: Modality Separation and Semantic Coherence}

Guided by these insights, we introduce \textbf{Lychee-FD}, a native end-to-end framework designed to resolve modality interference. Rather than circumventing these optimization bottlenecks via high-latency Thinker-Talker models (as illustrated in Figure~\ref{fig:architecture}), our architecture fundamentally tackles them via two targeted innovations: a \textbf{Hierarchical Parameter Separation} strategy to disentangle conflicting optimization directions, and a \textbf{Semantic Alignment Channel} to counteract semantic dilution and enforce robust knowledge retention.

\paragraph{Shared Backbone Foundation.}
We choose StepAudio-2-mini~\citep{DBLP:journals/corr/abs-2507-16632} as our half-duplex backbone due to its public availability. We employ the Whisper-v3-large encoder for input audio processing.
For acoustic output, we adopt the CosyVoice2 tokenizer to convert audio into discrete speech tokens at a 25Hz frame rate. 
Crucially, to ensure precise temporal alignment for full-duplex interaction, we utilize a 25Hz frame rate, distinct from the setting adopted by Step-Audio-2. 

\paragraph{Hierarchical Parameter Separation.}
Guided by the observation that acoustic and semantic gradients become orthogonal in deeper layers, we design a hierarchical Transformer architecture. We retain a unified Transformer backbone for the shallow layers to leverage shared low-level feature processing. Formally, given the input embedding sequence $\mathbf{E} \in \mathbb{R}^{N \times d}$, the shared representation $\mathbf{H}_{\text{shared}}$ is mathematically formulated as follows:
\begin{equation}
    \mathbf{H}_{\text{shared}} = \mathcal{F}_{\text{shared}}(\mathbf{E}; \theta_{\text{shared}}),
\end{equation}
where $\mathcal{F}_{\text{shared}}$ denotes the stack of shared Transformer layers parameterized by $\theta_{\text{shared}}$.

In the deeper layers, we physically disentangle the parameters into three specialized heads: the \textit{Semantic Head} for text generation, the \textit{Acoustic Head} for speech synthesis, and the \textit{Control Head} for interaction management (e.g., stop/start signals). These heads operate in parallel as follows:
\begin{equation}
    \mathbf{O}^{m} = \mathcal{F}_{\text{head}}^{m}(\mathbf{H}_{\text{shared}}; \theta_{m}), \quad m \in \{T, A, C\}
\end{equation}
where $m$ represents the modality (Text, Acoustic, Control), and $\mathbf{O}^{m}$ denotes the output logits for each head. 
The total cross-entropy loss $\mathcal{L}$ is computed as the summation of the next-token prediction losses for each specific head as follows:
\begin{equation}
    \mathcal{L} = - \sum_{m \in \{T, A, C\}} \sum_{t} \log P(y_t^m | y_{<t}, \mathbf{E}; \theta),
\end{equation}
where $y_t^m$ denotes the ground-truth token for modality $m$ at step $t$. This hierarchical split effectively isolates the conflicting optimization objectives, allowing the model to articulate high-fidelity acoustic responses without corrupting its underlying semantic modeling.

\paragraph{Semantic Alignment Channel.}
To counter the semantic dilution caused by sparse alignment in speech-native tasks, we incorporate a semantic alignment channel to generate coherent internal monologues. During training, these monologues serve as a semantic anchor, maintaining high-magnitude gradient flow for the language modeling objective. Specifically, we organize the parallel generation streams as follows:
\begin{subequations}
\begin{align*}
    Y^{\text{T}} &= [t_{1}, t_{2}, \cdots, t_{n}, \texttt{<EOT>}, \texttt{<pad>}, \cdots, \texttt{<pad>}], \\
    Y^{\text{A}} &= [a_{1}, a_{2}, \cdots, a_{n}, a_{n+1}, a_{n+2}, \cdots, \texttt{<EOS>}], \\
    Y^{\text{C}} &= [\texttt{<Start>}, c_{1}, \cdots, \cdots, \cdots, \texttt{<Stop>}],
\end{align*}
\end{subequations}
where $Y^{\text{C}}$ employs special tokens (e.g., $\texttt{<Start>}$ and $\texttt{<Stop>}$) to manage the onset and offset of the model's speech. By explicitly modeling the text channel alongside the acoustic channel, we ensure high-magnitude gradient flow for the language modeling objective, thereby preserving robust knowledge retention.

\begin{table*}[t]
\scriptsize
\centering
\setlength{\tabcolsep}{4pt} 
\label{tab:qa_benchmarks_expanded}
\begin{tabular}{l c cc cc cc cc c}
\toprule
\multirow{2}{*}{\textbf{Model}} & \multirow{2}{*}{\textbf{Type}} & 
\multicolumn{2}{c}{\textbf{LlamaQ}} & 
\multicolumn{2}{c}{\textbf{WebQ}} & 
\multicolumn{2}{c}{\textbf{TriviaQA}} & 
\multicolumn{2}{c}{\textbf{Avg.}} &
\multirow{2}{*}{\textbf{Takeover Rate}} \\
\cmidrule(lr){3-4} \cmidrule(lr){5-6} \cmidrule(lr){7-8} \cmidrule(lr){9-10}
 & & $S{\rightarrow}T$ & $S{\rightarrow}S$ 
 & $S{\rightarrow}T$  & $S{\rightarrow}S$  
 & $S{\rightarrow}T$  & $S{\rightarrow}S$  
 & $S{\rightarrow}T$  & $S{\rightarrow}S$ & \\
\midrule

 Freeze-Omni     & System-level & 71.3 & 50.7 & \underline{38.3} & 25.8 & 24.3 & 23.9 & 44.6 & 33.4 & 99.6 \\
 VITA 1.5        & System-level & \textbf{75.7} & 51.0 & \textbf{41.8} & \underline{29.2} & \underline{35.0} & 26.0 & \underline{50.8} & 35.4 & \textbf{100} \\
\midrule
 dGSLM           & Native   & -- & 1.3  & -- & 0.2  & -- & 0.4  & -- & 0.6  & \textbf{100} \\
 FLM-audio       & Native   & 41.3 & 36.7 & 15.6 & 14.5 & 10.5 & 10.4 & 22.4 & 20.5 & 99.5 \\
 Moshi           & Native   & 62.3 & 54.7 & 25.3 & 19.6 & 19.1 & 17.4 & 35.5 & 30.5 & 93.8 \\
 SALMONN-omni$^\ast$   & Native   & 67.0 & 61.7 & 33.7 & 28.1 & 32.9 & 24.2 & 44.5 & 38.0 & \underline{99.9} \\
 Fun-Audio-Chat  & Native   & 72.3 & \underline{64.3} & 26.2 & 24.4 & 29.6 & \underline{27.7} & 42.7 & \underline{38.8} & \underline{99.9} \\
\midrule
 StepAudio-2-mini & Half-duplex & 74.7 & 62.0 & 39.9 & 30.8 & 39.5 & 29.8 & 51.3 & 40.9 & -- \\
 \addlinespace[2pt]
\hdashline 
\addlinespace[2pt]
 Lychee-FD (Ours) & Native & \underline{73.7} & \textbf{65.3} & \underline{38.3} & \textbf{33.9} & \textbf{42.5} & \textbf{39.4} & \textbf{51.5} & \textbf{46.2} & \textbf{100} \\
 \quad w/o $Sem$-$Channel$  &            & 69.3 & 61.0 & 34.1 & 31.5 & 34.2 & 30.1 & 45.9 & 40.8 & 99.6 \\
 \quad w/o $Param$-$Sep$   &            & 67.0 & 36.0 & 34.6 & 22.5 & 36.6 & 24.2 & 46.1 & 27.6 & 98.5 \\
\bottomrule
\end{tabular}
\caption{Performance comparison on spoken question answering benchmarks. We report accuracy (Acc) in both speech-to-text ($S{\rightarrow}T$) and speech-to-speech ($S{\rightarrow}S$) settings. We also report average takeover rate (TOR) across three benchmarks. \textbf{Bold} denotes best results and \underline{underlined} denotes second best. $^\ast$ denotes our implementation.}
\label{tab:spoken_qa}
\vspace{-10pt}
\end{table*}

\section{Experiments}

\subsection{Baselines}

To ensure a comprehensive evaluation, we compare our proposed method against a diverse set of representative and competitive full-duplex SLMs. These baselines cover the primary architectural paradigms currently explored in the field:
\paragraph{System-Level Full-Duplex Models} include Freeze-Omni~\citep{DBLP:conf/icml/WangLFZS000M25} and VITA-1.5~\citep{DBLP:journals/corr/abs-2501-01957}, which integrate an external VAD module to manage the dialogue state of a standard half-duplex SLM.

\paragraph{Native Full-Duplex Models} include dGSLM \citep{DBLP:journals/tacl/NguyenKCAHETASM23}, FLM-Audio~\citep{DBLP:journals/corr/abs-2509-02521}, Moshi~\citep{DBLP:journals/corr/abs-2410-00037} and Fun-Audio-Chat~\citep{chen2025funaudiochattechnicalreport}. Specifically, Fun-Audio-Chat adopts a Thinker-Talker architecture, while the others utilize a CDM architecture.

\subsection{Training Data Construction}

Given the scarcity of open-source full-duplex datasets, we developed an automated pipeline to synthesize high-quality training data covering three key interaction behaviors: Interruptions, User Backchannels, and AI Backchannels. We employed multiple agents to simulate realistic User-Assistant dialogues, injecting rule-based constraints to trigger diverse interruption types (e.g., topic switching, follow-up queries) and natural backchannels. To ensure acoustic robustness, we synthesized speech from the transcripts via CosyVoice 2~\cite{DBLP:journals/corr/abs-2412-10117}, coupled with 80K predefined voice prompts for zero-shot cloning. After rigorous filtering to remove samples with logical inconsistencies or low audio quality, we curated a final dataset of approximately 140K full-duplex dialogue instances, providing a diverse and reliable foundation for our experiments. We provide more details of our data pipeline in Appendix~\ref{sec:appendix_synthesis}.

\subsection{Implementation Details}

We optimize our model using AdamW~\citep{DBLP:conf/iclr/LoshchilovH19} with a cosine learning rate scheduler. All experiments are conducted on 8 NVIDIA H20 GPUs, with a global batch size of 32 and a learning rate of 3e-6. We set the warmup ratio to 0.1 and train for 1 epoch, which takes approximately 16 hours. For inference, we employ greedy sampling for both text and speech token generation. We evaluate our model with three random seeds and report their average performance.
Regarding the hierarchical architecture configuration, unless otherwise specified, we utilize a shared backbone of 24 Transformer layers. The specialized heads are configured with 4 layers for the text channel, 4 layers for the speech channel, and 2 layers for the control channel. This yields approximately 10B total parameters, a choice supported by the marginal gain analysis in Appendix~\ref{app:layer_ablation}.

\subsection{Spoken Question Answering}

\paragraph{Metrics.} To evaluate the speech intelligence capabilities of our model, we follow previous work~\citep{DBLP:journals/corr/abs-2410-00037} and utilize three standard spoken question answering benchmarks: LlamaQ, WebQ, and TriviaQA. We report the accuracy (Acc) under both speech-to-text ($S \rightarrow T$) and speech-to-speech ($S \rightarrow S$) settings. For speech-to-speech setting, we leverage Whisper-large-v3~\citep{DBLP:conf/icml/RadfordKXBMS23} to obtain the transcription of generated speech. Additionally, we report the takeover rate (TOR) across three benchmarks to quantify the frequency of model responses, serving as an indicator of the model's turn-taking behavior.

\paragraph{Result.} 
As presented in Table~\ref{tab:spoken_qa}, Lychee-FD demonstrates superior spoken question answering capabilities. It achieves the highest average accuracy across both speech-to-text ($S \rightarrow T$) and speech-to-speech ($S \rightarrow S$) settings. 
Compared to the previous SOTA native full-duplex model, Fun-Audio-Chat, Lychee-FD delivers a substantial improvement of 7.4\% in $S \rightarrow S$ accuracy and 8.8\% in $S \rightarrow T$ accuracy.
Even when compared to system-level pipelines like VITA-1.5, our end-to-end approach demonstrates superior reasoning capabilities (10.8\% in $S \rightarrow S$ and 0.7\% in $S \rightarrow T$), validating the effectiveness of our framework. Furthermore, Lychee-FD maintains a perfect TOR of 100\%, confirming that this exceptional knowledge retention is achieved without compromising seamless, real-time interaction stability.

\begin{table*}[t]
\centering
\small
\resizebox{\linewidth}{!}{
\begin{tabular}{l cccccc cccccc cccc}
\toprule
\multirow{3}{*}{\textbf{Model}} & 
\multicolumn{6}{c}{\textbf{FDBench}} & 
\multicolumn{6}{c}{\textbf{FullDuplexBench 1.0}} & 
\multicolumn{4}{c}{\textbf{FullDuplexBench 1.5}} \\
\cmidrule(lr){2-7} \cmidrule(lr){8-13} \cmidrule(lr){14-17}

& SRR$\uparrow$ & SIR$\uparrow$ & EIR$\downarrow$ & SRIR$\uparrow$ & FSED.$\downarrow$ & IRD$\downarrow$ & I-TOR$\uparrow$ & B-Freq$\uparrow$ & B-TOR$\downarrow$ & T-TOR$\uparrow$ & P-TOR$\downarrow$ & Stop$\downarrow$ 
& IRR$\uparrow$ & BRR$\uparrow$ & Stop$\downarrow$ & Lat.$\downarrow$ \\
\midrule
dGSLM           & -- & -- & -- & -- & -- & -- & 91.7 & 1.5 & 69.1 & \underline{97.5} & 93.5 & 2523 & -- & -- & -- & -- \\
Freeze-Omni     & 12.9 & 57.2 & 25.7 & 29.5 & \underline{667}  & 5413  & 77.5 & 0.1 & 63.6 & 33.6 & 46.3 & 1380 & 27.0 & 63.0 & \underline{660} & 2066 \\
VITA 1.5        & 21.0 & 46.1 & 16.3 & \underline{78.3} & 3036 & 9925  & \textbf{99.5} & 2.5 & 81.8 & 58.8 & 88.8 & 1523 & 6.0 & 38.0 & 1222 & 2140 \\
FLM-audio       & 7.5 & 69.4 & \underline{1.0} & 0.9 & 989 & 3408  & 91.0 & 0.3 & 61.8 & 96.6 & 56.5 & 4579 & 10.0 & \underline{43.0} & 2439 & \underline{983} \\
Moshi           & \underline{41.4} & \underline{78.8} & 22.1 & 73.9 & 1895 & \underline{1421}  & 87.5 & \underline{5.1} & \underline{36.4} & 76.4 & 54.1 & \underline{885} & \underline{61.0} & 26.0 & 1071 & 3034 \\
\midrule
Lychee-FD (Ours)   & \textbf{86.3} & \textbf{99.7} & \textbf{0.4} & \textbf{95.8} & \textbf{637} & \textbf{1210} & \underline{94.5} & \textbf{14.6} & \textbf{23.4} & \textbf{98.3}  & \textbf{10.0} & \textbf{840} & \textbf{78.0} & \textbf{69.0} & \textbf{570} & \textbf{826} \\
\bottomrule
\end{tabular}
}
\caption{Performance comparison of full-duplex interaction capabilities and efficiency. Despite interaction behavior metrics, we also report efficiency metrics of each benchmark (FSED, IRD, Lat. Stop.), measured in milliseconds (ms).  $\uparrow$ indicates higher is better. \textbf{Bold} denotes best results and \underline{underlined} denotes second best.}
\label{tab:full_duplex_chatting} 
\vspace{-10pt}
\end{table*}

\paragraph{Ablation.} 
To systematically investigate the individual contributions of our two proposed innovations, we conducted ablation studies on two variants: \textit{w/o Sem-Channel}, which replaces the semantic alignment channel with sparse time-aligned text, and \textit{w/o Param-Sep}, which removes the hierarchical parameter separation to employ a fully shared architecture. 
As shown in Table~\ref{tab:spoken_qa}, when applying the time-aligned text, we observe a significant performance decline in both $S \rightarrow T$ (5.6\%) and $S \rightarrow S$ (5.4\%) settings. This parallel degradation confirms our hypothesis regarding semantic dilution: the sparse supervision provided by time-aligned text fails to sustain robust linguistic modeling, which in turn causes knowledge degradation. 
In contrast, the \textit{w/o Param-Sep} setting reveals severe modality interference. While its text generation capability remains relatively stable, its speech accuracy suffers a catastrophic drop to 27.6\%. This disparity validates our optimization dynamics analysis (as illustrated in Figure~\ref{fig:motivation}): without physically disentangling the shared parameters, the optimization landscape becomes dominated by semantic modeling, effectively suppressing the learning of acoustic features in deep layers.

\paragraph{Discussion.} 
We highlight two critical phenomena that distinguish Lychee-FD from existing paradigms. 
First, our model not only recovers the performance of its half-duplex backbone (StepAudio-2-mini) but surpasses it in both $S \rightarrow T$ (+0.2\%) and $S \rightarrow S$ (+5.3\%) settings. This performance gain, achieved without increasing model depth, strongly validates that our framework realizes robust knowledge retention while maintaining the inference efficiency of the native end-to-end model. This underscores the pivotal role of explicit semantic modeling in driving model intelligence, offering valuable insights for future human-machine interaction systems.
Second, Lychee-FD achieves the smallest modality gap among all native CDM models (e.g., FLM-Audio and Moshi) and remains competitive with Thinker-Talker architectures, without requiring intricate multi-stage training. This demonstrates that by physically decoupling semantic and acoustic modeling, Lychee-FD effectively resolves modality interference. Consequently, our approach allows the model to articulate nature acoustic responses without corrupting its underlying semantic logic, providing a streamlined solution to the efficiency-intelligence trade-off.

\subsection{Full-duplex Chatting}

\paragraph{Metrics.} For the Full-duplex Chatting, we select three mainstream benchmarks: \textbf{FDBench}~\citep{DBLP:conf/interspeech/PengCNMN0C25}, \textbf{FullDuplexBench 1.0}, and \textbf{FullDuplexBench 1.5}~\citep{DBLP:journals/corr/abs-2503-04721}. We follow the recommended settings of these benchmarks for evaluating both baselines and our method. Specifically, FDBench assesses turn-taking and interruption behaviors using Success-Replies Rate (SRR), Success-Interrupts Rate (SIR), Early-Interrupts Rate (EIR), and Success-Replies-to-Interrupts Rate (SRIR), alongside timing metrics such as First-Speech-Emit-Delay (FSED) and Interrupt-Response-Delay (IRD). 
FullDuplexBench 1.0 consists of four subsets: interruption (I), assistant backchannel (B), turn-taking (T), and user pause (P). In addition to the takeover rate (TOR) and Interruption stop delay (Stop.), we also report the frequency of generated backchannels during user speech (B-Freq). 
Finally, FullDuplexBench 1.5 utilizes GPT-4o-1124 to classify model responses to user interruptions and backchannels into four categories (Response, Resume, Uncertain, or Unknown), reporting the Interruption-Response Rate (IRR), Backchannel-Resume Rate (BRR), as well as the interruption stop delay and response latency.

\paragraph{Result.} 
As presented in Table~\ref{tab:full_duplex_chatting}, Lychee-FD achieves state-of-the-art performance across 10 of the 11 evaluated interaction metrics, demonstrating superior capability in managing complex conversational dynamics, including interruption, backchanneling, dynamic turn-taking, and pause handling.
While VITA-1.5 exhibits a marginally higher I-TOR on FullDuplexBench 1.0, its consistently high B-TOR and P-TOR reveal a tendency towards aggressive, indiscriminate speech rather than intelligent turn-taking. In contrast, Lychee-FD maintains a balanced interaction profile, effectively distinguishing between user pauses and interruptions. 
Notably, on the challenging FullDuplexBench 1.5, our model delivers a substantial 28.5\% average improvement over system-level baselines (e.g., Freeze-Omni) that rely on external VAD. 
Crucially, this empirical breakthrough directly validates our core scientific insight: by resolving the inherent gradient conflicts that plague fully shared architectures, Lychee-FD preserves robust semantic awareness during simultaneous listening and speaking. 
Consequently, this result not only proves that our model realizes truly natural, fluid, and immersive interaction, but also confirms that LLMs can intrinsically function as highly effective dialog managers. By eliminating handcrafted signal processing modules, our approach provides an efficient, practical solution to the core problem of full-duplex communication, substantially pushing forward the frontier of native end-to-end SLMs.

\paragraph{Latency.} 
We evaluate inference efficiency through two categories: first response latency (FSED, Lat.) and interruption response latency (IRD, Stop.).
Regarding the former, Lychee-FD achieves the lowest latency across all benchmarks. Since our hierarchical parameter separation strategy introduces no additional model depth, we can leverage standard pipeline parallelism for speedup, demonstrating the efficiency of our architectures. Further details about this real-time parallel inference are provided in Appendix~\ref{app:inference_system}.
For Interruption Response Latency, our model demonstrates an even greater advantage, achieving the lowest stop latency when meeting an interruption (e.g., 570ms Stop. on FullDuplexBench 1.5). It is worth noting that interruption latency is a function of both model processing speed and interaction accuracy---since a model must first correctly identify an interruption before it can stop generating. The superior performance confirms that our architecture incurs no computational overhead, and that its robust semantic awareness enables rapid, accurate reactions to user interventions.

\subsection{Speech Generation}






\noindent
\begin{minipage}[t]{0.48\textwidth}
    \centering
    \includegraphics[width=0.98\linewidth]{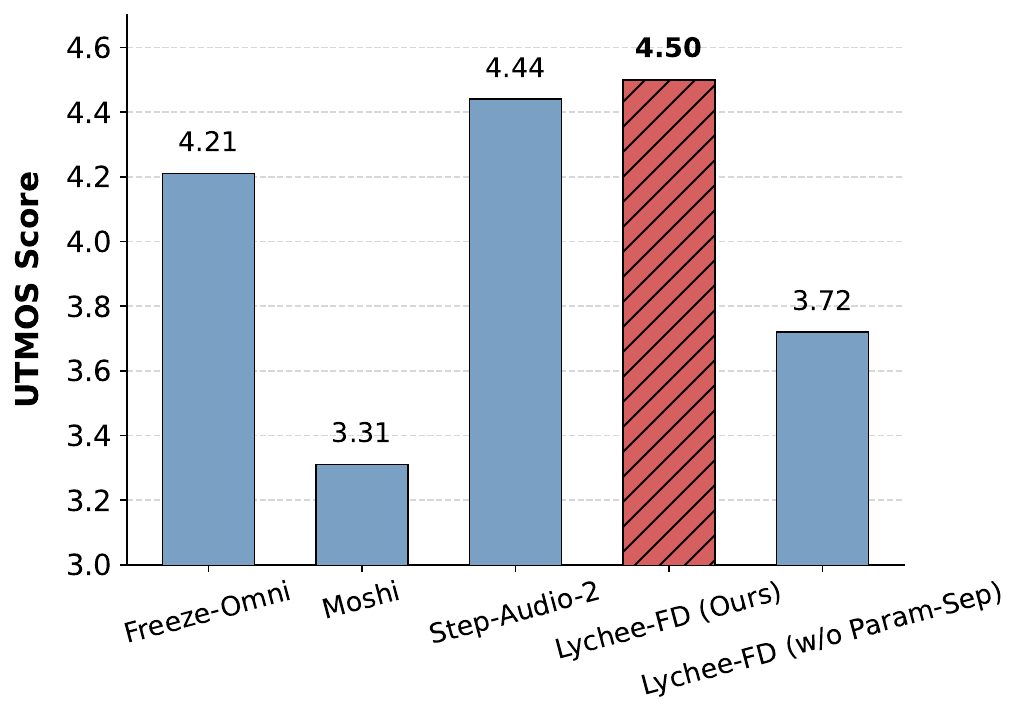}
    \captionof{figure}{Comparison of speech synthesis quality via UTMOS. Lychee-FD (\textit{w/o Param-Sep}) denotes the variant without hierarchical parameter separation strategy.}
    \label{fig:speech_quality}
\end{minipage}
\hfill
\begin{minipage}[t]{0.48\textwidth}
    \centering
    \includegraphics[width=0.98\linewidth]{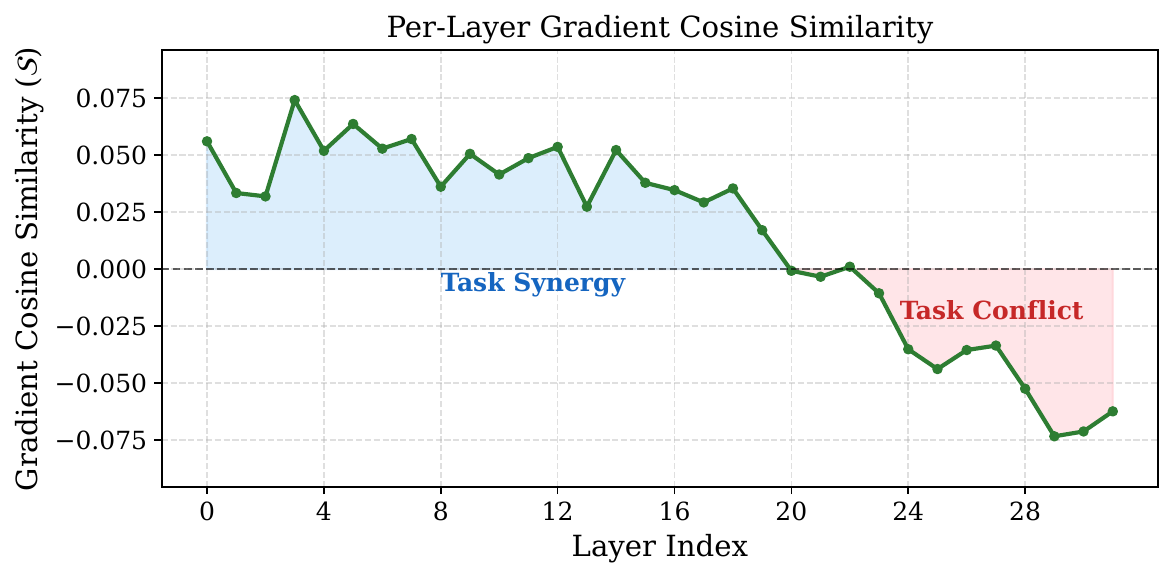}
    \captionof{figure}{Gradient Cosine Similarity on the Moshi architecture. The transition from task synergy in shallow layers to task conflict in deep layers confirms that modality interference is a universal phenomenon in fully shared end-to-end SLMs.}
    \label{fig:moshi_gradient}
\end{minipage}

\paragraph{Metrics.} To investigate the quality of the generated speech, we evaluate the content consistency and speech naturalness of the model on LlamaQ dataset. Specifically, we report the Word Error Rate (WER) between the generated text and the transcribed speech to measure content consistency. Additionally, we employ UTMOS~\citep{DBLP:conf/interspeech/SaekiXNKTS22}, a trained speech quality assessment model, to score the naturalness of the generated audio.

\paragraph{Result.} High-fidelity speech synthesis serves as the cornerstone of voice interaction, significantly elevating the immersive quality of full-duplex conversations. 
As illustrated in Figure~\ref{fig:speech_quality}, Lychee-FD achieves the highest UTMOS score of 4.50, surpassing Freeze-Omni, Moshi and even its half-duplex backbone. 
This result empirically validates that our proposed hierarchical parameter separation strategy effectively prevents acoustic modeling from semantic interference, thereby preserving fine-grained prosodic details that are often lost in shared architectures.
When the parameter separation is removed, we observe a significant decline in speech quality. This finding corroborates our hypothesis regarding modality interference from an acoustic perspective, demonstrating that forcing the acoustic head to share deep layers with text processing objectives inevitably degrades audio fidelity.
Ultimately, by physically disentangling the modeling pathways, Lychee-FD not only improves generation accuracy but also synthesizes speech with richer acoustic details, delivering a more natural and enjoyable conversational experience.

\section{Optimization Dynamics Analysis Extension}
\label{app:moshi_dynamics}

To verify the universality of our findings, we extend the gradient cosine similarity analysis to Moshi~\citep{DBLP:journals/corr/abs-2410-00037}, investigating whether deep-layer gradient conflicts are architecture-specific artifacts or a fundamental bottleneck across different FDSLMs.
Following our established methodology, we find that Moshi's optimization dynamics exhibit a strikingly similar trend (Figure~\ref{fig:moshi_gradient}). Shallow layers (0-19) maintain positive similarity ($\mathcal{S}^{(l)} > 0$), indicating task synergy. Conversely, as depth increases, the similarity sharply declines to distinctly negative values in deep layers (23-31). This highlights the optimization divergence between text and speech tasks.
This cross-architecture validation empirically confirms that inherent gradient conflicts between semantic and acoustic modeling are not isolated anomalies, but a universal bottleneck in fully shared paradigms. Consequently, it firmly substantiates the necessity and generalizability of our proposed Hierarchical Parameter Separation.

\section{Conclusion}

In this paper, we address a critical challenge and a long-standing goal for the speech and NLP communities: developing seamless, high-performance native full-duplex spoken language models. To the best of our knowledge, we are the first to uncover the fundamental root cause of modality interference. Through an in-depth analysis of model optimization dynamics, we reveal that the inability to simultaneously listen and speak stems from inherent gradient conflicts within a shared deep parameter space, which specifically manifest as optimization divergence and semantic dilution.
Inspired by these observations, we introduce Lychee-FD. Our framework elegantly resolves these bottlenecks by decoupling conflicting modalities via a hierarchical parameter separation strategy, coupled with a semantic alignment channel to enforce robust knowledge retention. Extensive experiments demonstrate that Lychee-FD significantly advances the state-of-the-art performance, successfully reconciling ultra-low latency interaction with deep speech intelligence. 
Ultimately, our work substantially pushes forward the frontier of full-duplex SLMs research, paving the way for the next generation of natural, fluid, and immersive human-machine interaction.

\section*{Limitations}

While Lychee-FD enables the development of a high-performance native full-duplex framework with highly responsive acoustic processing capabilities, deploying such systems in unconstrained, real-world open-mic scenarios remains a promising yet challenging next research frontier. At present, our model implements highly sensitive interruption detection and can effectively suspend speech output upon any user barge-in. However, discriminating between intentional user instructions and incidental background side-talk (as illustrated in Appendix~\ref{app:error_analysis}) remains a non-trivial challenge to address. This limitation is tentatively attributed primarily to the current data synthesis paradigm, which is largely centered on direct two-party interactions and lacks detailed intent annotations for complex multi-speaker scenarios. This issue points to a critical gap in data coverage within the field, rather than stemming from an architectural bottleneck of the hierarchical framework we propose. Moving forward, we hope our preliminary work can provide a modest impetus for subsequent research in the community. By constructing diverse open-mic datasets and integrating prosodic features for intent disambiguation, future studies may build upon this initial framework to develop truly context-aware and adaptive selective interruption mechanisms for next-generation SLMs.

\newpage

\bibliographystyle{lychee}
\bibliography{custom}

\newpage

\section{Layer Ablation}



\begin{wrapfigure}{r}{0.48\textwidth}
\vspace{-15pt}
\centering
\includegraphics[width=0.98\linewidth]{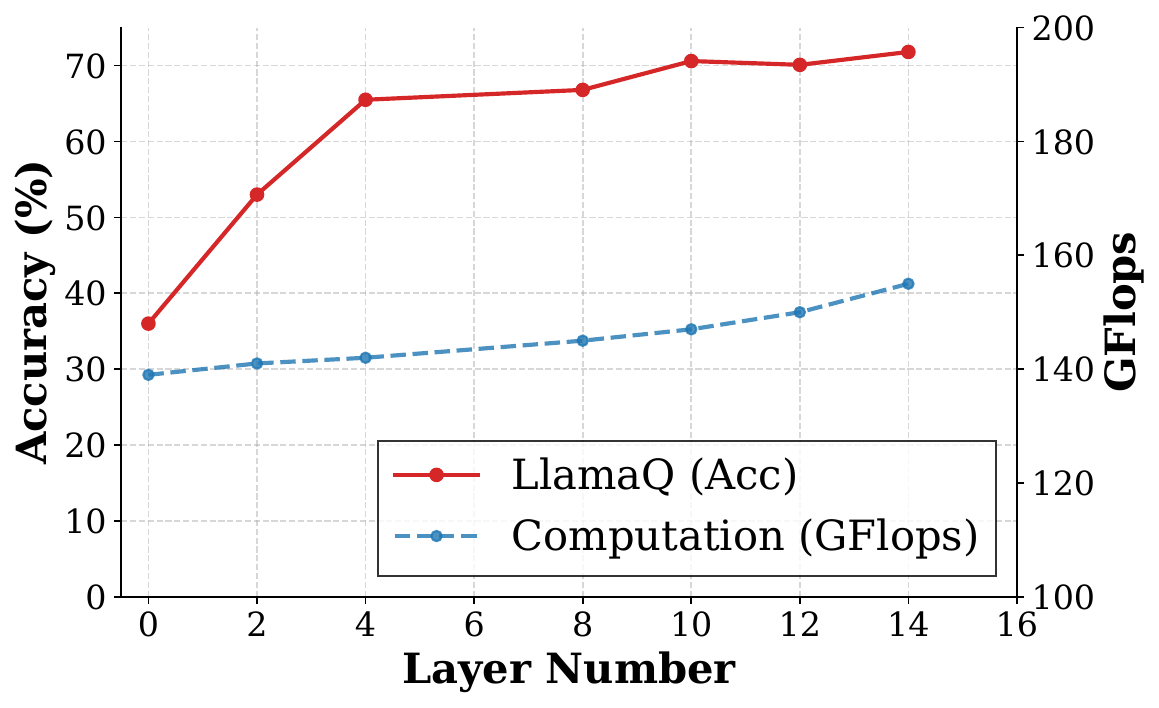}
\caption{Layer ablation study on model performance and computation cost. Accuracy on LlamaQ saturates after 4 separated layers, while computational overhead grows linearly. This indicates that 4 layers represent the optimal trade-off between performance gains and inference efficiency.}
\label{fig:layer_ablation}
\vspace{-15pt}
\end{wrapfigure}

To explore the impact of the number of separated layers, we conduct a layer ablation study on both model performance on LlamaQ and static inference cost for speech, as illustrated in Figure~\ref{fig:layer_ablation}. 
We observe that accuracy follows a steep upward trajectory in the initial stages, surging from 36.0 to 65.4 as the depth increases to 4 layers. This indicates that a relatively shallow separation is sufficient to resolve the primary modality conflicts.
However, beyond this point, the performance gain saturates, yielding diminishing returns, while the computational overhead continues to grow linearly. 
Consequently, we identify 4 layers as the optimal configuration. This choice represents the most favorable trade-off, securing the vast majority of the performance gain while minimizing the additional parameter budget, thereby ensuring high inference efficiency.

\vspace{20pt}

\label{app:layer_ablation}

\section{Taxonomy of Full-Duplex Scenarios}
\label{app:taxonomy}

To intuitively demonstrate the interaction quality of Lychee-FD, we move beyond isolated examples and analyze user interventions through two basic dimensions: the presence of voice activity and the user's actual intent to intervene. Since the default state (no voice activity and no intent) simply requires the model to continue listening or speaking, we focus on the remaining three representative scenarios: intentional interruption (voice with strong intent to intervene), passive backchannel (voice with intent to encourage continuation), and side-talk (voice with no intent towards the agent).

\begin{figure*}[ht!]
\centering
\includegraphics[width=\linewidth]{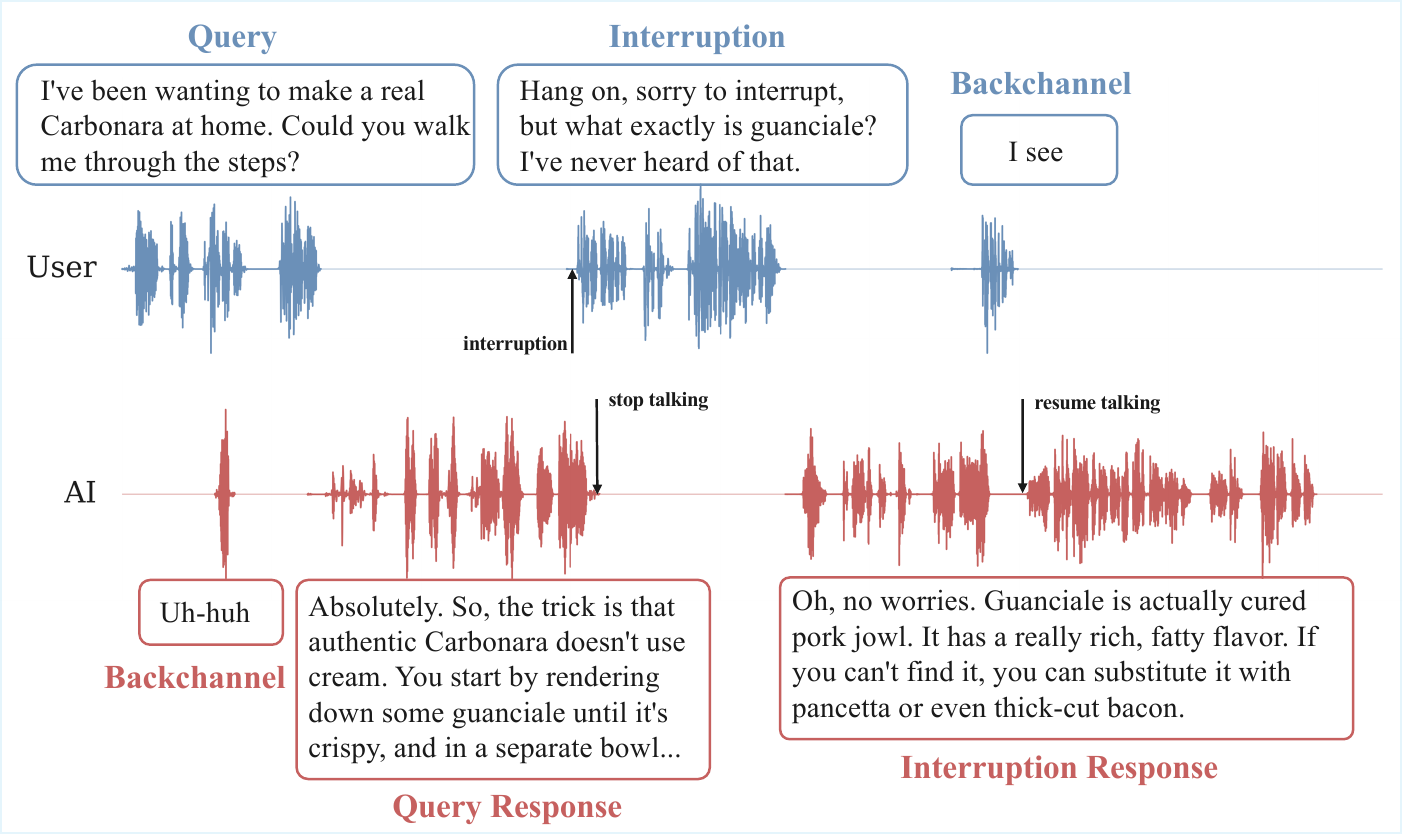}
\caption{A case study demonstrating Lychee-FD's capability in handling complex turn-taking dynamics. Lychee-FD successfully generates backchannels, halts immediately upon interruption, and provides a contextually accurate response to the user's specific query without misinterpreting passive feedback.}
\label{fig:case_study}
\vspace{-10pt}
\end{figure*}

\begin{figure*}[ht!]
\centering
\includegraphics[width=\linewidth]{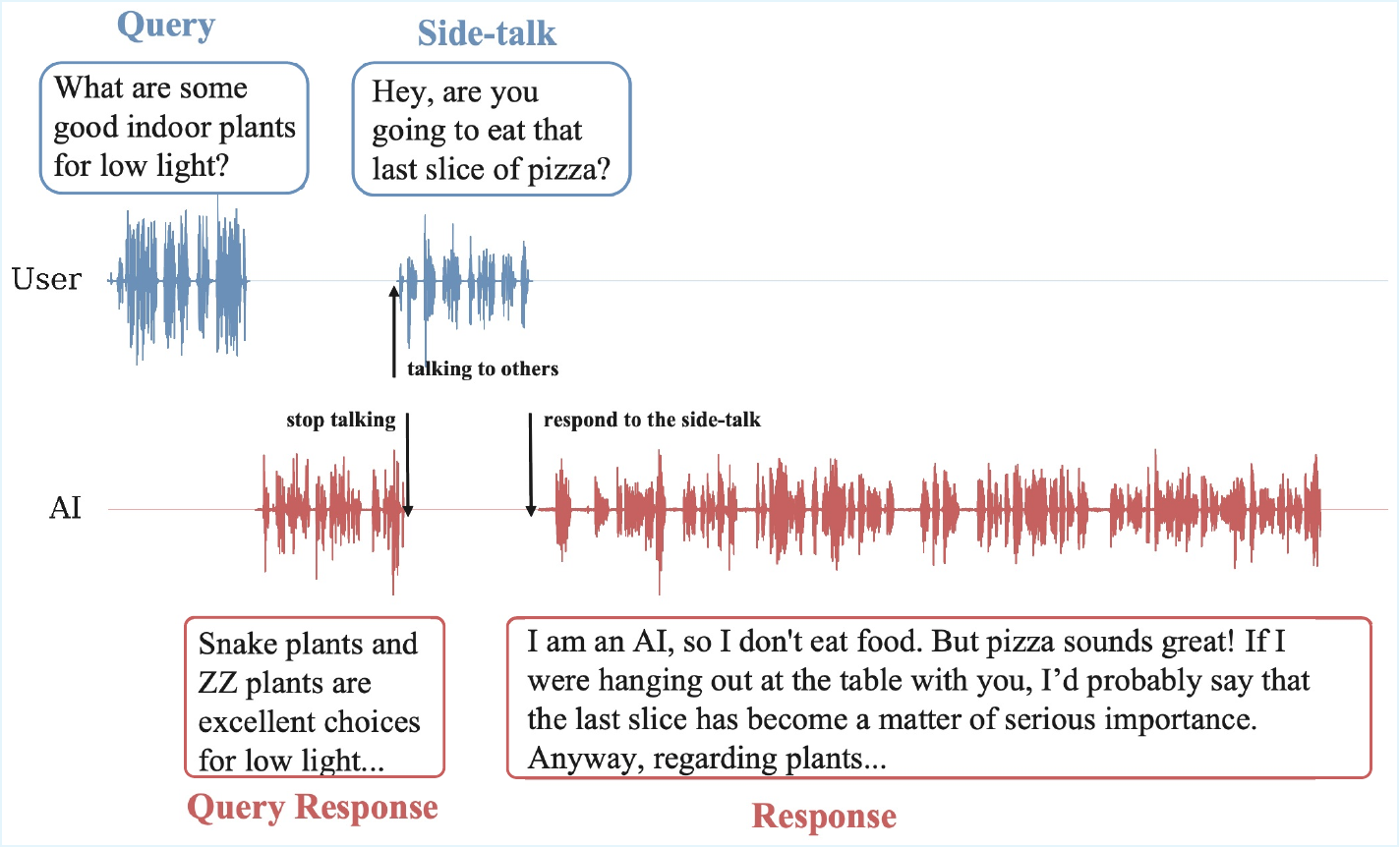}
\caption{An error analysis illustrating a common limitation in handling side-talk during full-duplex interaction. While the model correctly detects voice activity and halts its speech, it fails to recognize that the user's utterance is directed at a third party, highlighting a key area for future community research.}
\label{fig:error_analysis}
\vspace{-10pt}
\end{figure*}

\subsection{Case Study: Interruption and Backchannel}

In Figure~\ref{fig:case_study}, we present a real-world conversation sample that illustrates the first two scenarios. As the model begins explaining a recipe, it first employs a natural backchannel (``Uh-huh'') to acknowledge the user's start. Crucially, the model exhibits two key capabilities when facing different user interventions:
\textbf{(1) Intentional Interruption:} When the user interrupts with a specific clarification question (``what exactly is guanciale?''), Lychee-FD halts its speech output almost instantaneously, avoiding the awkward talking-over phenomenon common in half-duplex systems.
\textbf{(2) Passive Backchannel:} Later, as the model explains the definition of guanciale, the user interjects with a short backchannel (``I see''). Here, Lychee-FD demonstrates precise intent understanding. Instead of misinterpreting this acoustic signal as a barge-in command---a common failure in fully-shared models due to semantic dilution---the model correctly identifies it as a passive signal of agreement. Benefiting from our proposed Semantic Alignment Channel, which preserves robust language modeling during simultaneous listening, Lychee-FD seamlessly resumes its explanation without unnecessary pauses or topic fragmentation. 
This interaction confirms that our Lychee-FD effectively maintains the model's language capabilities even during rapid turn-switching, enabling a fluid, seamless, and truly natural conversational experience.

\subsection{Error Analysis: Side-Talk} 
\label{app:error_analysis}

Despite the strong performance in standard interactions, we observe certain limitations in handling complex acoustic environments, particularly regarding side-talk. In a native full-duplex setting, the model continuously processes incoming audio. As shown in Figure~\ref{fig:error_analysis}, when the user briefly speaks to a third party in the background (e.g., asking a roommate about food), the model incorrectly interprets this background conversation as a direct interruption. Consequently, it halts its current explanation and attempts to respond to the irrelevant query.

This error indicates that while Lychee-FD is highly responsive to voice activity, its ability to distinguish between user-to-agent commands and user-to-human side-talk remains constrained. We argue that this is not an architectural limitation of our framework, but rather a common challenge for current native SLMs, largely due to the lack of multi-speaker intent annotations in existing data synthesis paradigms. By highlighting this scenario, we hope to provide a blueprint for the community. Future work will focus on integrating intent detection or utilizing prosodic cues to improve the model's robustness in open-mic, multi-speaker environments.

\section{Global Gradient Influence Analysis}

\label{app:gradient_influence}

\begin{figure}[htbp]
    \centering
    \includegraphics[width=\linewidth]{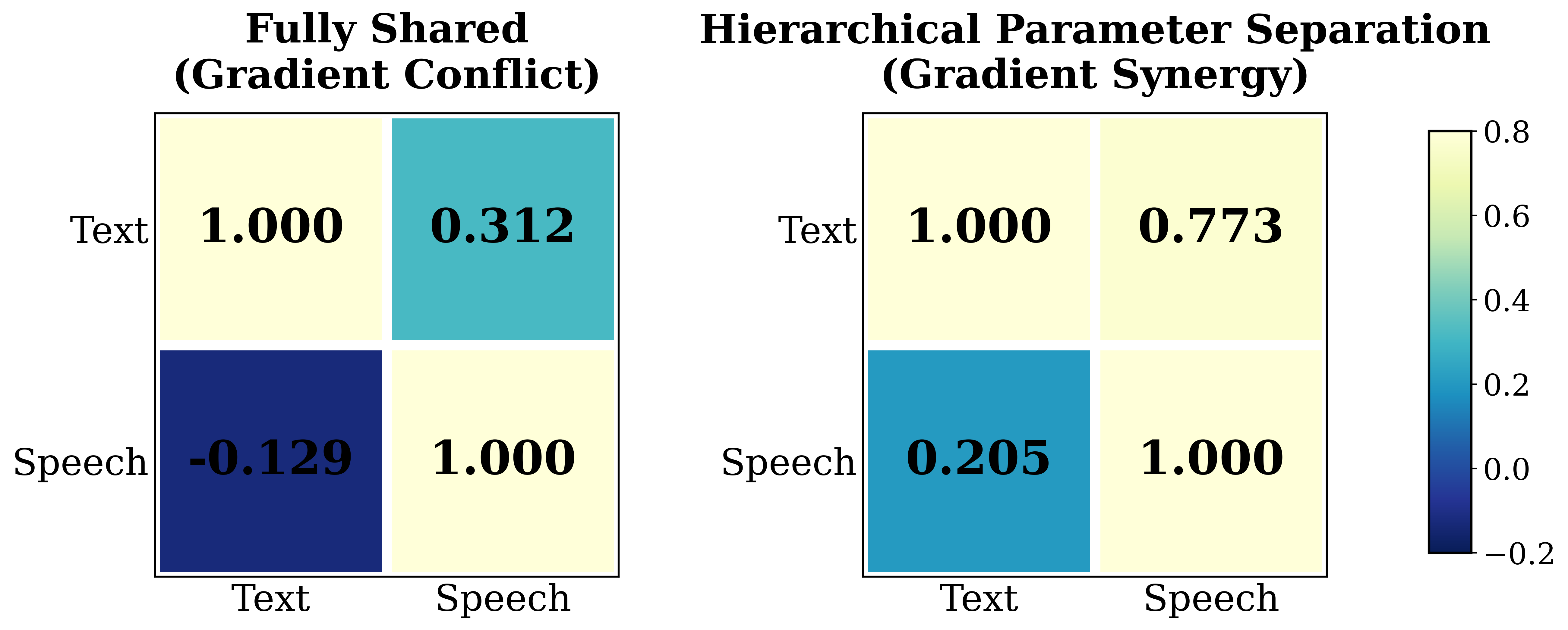} 
    \caption{\textbf{Global gradient influence scores among Text and Speech tasks.} Left: The fully shared baseline suffers from severe destructive interference between semantic and acoustic modeling (negative scores). Right: Our Hierarchical Parameter Separation not only eliminates this conflict but also fosters constructive synergy (positive scores) between modalities.}
    \label{fig:gradient_infulence}
\end{figure}

To further validate the effectiveness of our proposed Hierarchical Parameter Separation, we investigate modality interference from a causal perspective. Inspired by~\citet{DBLP:journals/corr/abs-2601-02967}, we quantify how updating parameters for one task causally affects the performance of another. 
Specifically, we calculate the normalized global Influence Score $I_{m \rightarrow i}$ of task $m$ on task $i$ ($m, i \in \{T, A\}$): 
\begin{equation}
    I_{m \rightarrow i} = \frac{\mathcal{L}_i(\theta) - \mathcal{L}_i(\theta - \eta \mathbf{g}_m)}{\mathcal{L}_i(\theta) - \mathcal{L}_i(\theta - \eta \mathbf{g}_i)},
\end{equation}
where $\theta$ denotes the global model parameters, and $\mathbf{g}_m = \nabla_\theta \mathcal{L}_m$ is the gradient derived solely from the objective of task $m$. The denominator represents the loss reduction when task $i$ is updated using its own gradient, serving as a normalization factor. A negative score ($I_{m \rightarrow i} < 0$) indicates destructive interference, meaning that optimizing task $m$ degrades the performance of task $i$. Conversely, a positive score ($I_{m \rightarrow i} > 0$) implies constructive synergy.

Figure~\ref{fig:gradient_infulence} visualizes the influence scores between the text and speech tasks. In the fully shared baseline (left), the update of the text task exerts a negative influence on the speech task (-0.129). This empirically confirms our hypothesis: forcing conflicting modalities to update within a shared parameter space causes destructive interference, where acoustic learning actively corrupts semantic modeling. In contrast, under our Hierarchical Parameter Separation (right), this destructive interference is entirely resolved. Notably, the influence score of text on speech shifts to a positive value (0.205). Furthermore, the constructive synergy from speech to text is significantly amplified (from 0.312 to 0.773). This demonstrates that our architecture effectively disentangles conflicting optimization dynamics, allowing the model to acquire robust speech capabilities while mutually reinforcing its language intelligence.

\section{Real-Time Inference Algorithm}
\label{app:inference_system}

Deploying Lychee-FD for real-time full-duplex interaction presents a unique system-level challenge. Standard LLM inference engines, such as vLLM, are fundamentally designed around a linear layer topology ($L_1 \rightarrow L_2 \rightarrow \dots \rightarrow L_D$) and a single autoregressive output stream. However, our proposed hierarchical architecture requires the simultaneous generation of text, audio, and control signals. Forcing this multi-head architecture into a single GPU or a linear pipeline would result in the sequential execution of the specialized heads, introducing severe latency bottlenecks that disrupt the fluidity of real-time conversation.

To bridge the gap between theoretical modality decoupling and practical deployment, we introduce \textbf{Directed Acyclic Graph Pipeline Parallelism (DAG-PP)} by customizing the vLLM engine. Instead of linear tensor passing, we implement a 1-to-N tensor broadcast mechanism at the branching point of the shared backbone. As illustrated in Algorithm~\ref{alg:three_branch_decode}, the intermediate hidden states ($\mathbf{H}_{\mathrm{shared}}$) are broadcasted via NCCL to multiple GPUs. This allows the Semantic, Acoustic, and Control heads to execute strictly in parallel across different devices. Crucially, this parallel execution ensures that the effective model depth on the critical path remains unchanged compared to the half-duplex backbone, \textbf{significantly reducing the inference latency bottleneck of multi-head architectures.} A distributed synchronization barrier is then employed to collect the multi-stream logits for synchronous sampling before the next autoregressive step.

This algorithm-system co-design demonstrates that our hierarchical parameter separation not only resolves gradient conflicts during training but also unlocks strict physical parallelism during inference. By hiding the computational overhead of multi-modal generation behind parallel execution, Lychee-FD achieves the state-of-the-art interaction intelligence while strictly adhering to the ultra-low latency constraints of full-duplex spoken dialogues.

\begin{algorithm}[t]
\caption{Directed Acyclic Graph Pipeline Parallelism (DAG-PP)}
\label{alg:three_branch_decode}
\small
\SetKwBlock{Parallel}{parallel for}{end}

\KwIn{Text tokens $\mathbf{X}^{T}$, Acoustic tokens $\mathbf{X}^{A}$, Control tokens $\mathbf{X}^{C}$, User speech $\mathbf{U}$}
\KwOut{Text logits $\mathbf{O}^{T}$, Acoustic logits $\mathbf{O}^{A}$, Control logits $\mathbf{O}^{C}$}

\BlankLine
\tcc{Initial Embedding (GPU 0)}
$\mathbf{E}^{m} \leftarrow \mathrm{Embedding}_{m}(\mathbf{X}^{m}), \quad \forall m \in \lbrace T,A,C \rbrace$\;
$\mathbf{E}^{U} \leftarrow \mathrm{AudioEncoder}(\mathbf{U})$\;
$\mathbf{E} \leftarrow \mathbf{E}^{T} + \mathbf{E}^{A} + \mathbf{E}^{C} + \mathbf{E}^{U}$\;

\BlankLine
\tcc{Shared Backbone Execution (GPU 0)}
$\mathbf{H}^{(0)} \leftarrow \mathbf{E}$\;
\For{$l \leftarrow 1$ \KwTo $L_{\mathrm{shared}}$}{
    $\mathbf{H}^{(l)} \leftarrow \mathcal{F}^{\mathrm{shared}}_l(\mathbf{H}^{(l-1)})$\;
}

\BlankLine
\tcc{DAG Branching}
\textbf{NCCL\_Broadcast}($\mathbf{H}_{\mathrm{shared}} \rightarrow \mathrm{GPU}_{T}, \mathrm{GPU}_{A}, \mathrm{GPU}_{C}$)\;

\BlankLine
\tcc{Specialized Heads Execution}
\Parallel{$m \in \lbrace T, A, C \rbrace$ \textbf{on} $\mathrm{GPU}_{m}$}{
    $\mathbf{H}^{m} \leftarrow \mathbf{H}_{\mathrm{shared}}$\;
    \For{$k \leftarrow 1$ \KwTo $L_{m}$}{
        $\mathbf{H}^{m} \leftarrow \mathcal{F}^{m}_k(\mathbf{H}^{m})$\;
    }
}

\BlankLine
\tcc{Distributed Synchronization}
\textbf{Barrier\_Synchronize}()\;

\Return $(\mathbf{O}^{T}, \mathbf{O}^{A}, \mathbf{O}^{C})$\;
\end{algorithm}

\section{Data Synthesis Pipeline Details}
\label{sec:appendix_synthesis}

To address the scarcity of full-duplex interaction data, we developed an automated data synthesis pipeline. This pipeline orchestrates interactions between a \textbf{User Agent} and an \textbf{Assistant Agent}, review by a \textbf{Reviewer Agent}. The process explicitly models complex conversational behaviors including interruptions and backchannels.

\subsection{Agent Architecture}
\begin{itemize}
    \item \textbf{User Agent} is initialized with a specific \textit{Persona} (randomly sampled from a pool of diverse profiles) and a \textit{Speaking Style} (sampled from 19 distinct styles such as Concise and logical, Impatient, Humorous and witty). The agent is instructed to act authentically rather than helpfully.
    
    \item \textbf{Assistant Agent} generates responses based on the conversation history to simulate a realistic AI assistant.
    
    \item \textbf{Reviewer Agent} evaluates dialogue turns based on persona adherence, event execution quality, and logical flow.
\end{itemize}

\subsection{Interaction Behavior Modeling}

\paragraph{Interruption Generation.}
We implemented a two-stage mechanism to generate naturalistic interruptions. Random interruptions are scheduled between turns 2 and 4.
\begin{enumerate}
    \item \textbf{Planning Phase:} The User Agent analyzes the Assistant's current response context to determine a valid \textit{Interruption Motivation} (Correction, Deeper Inquiry, Topic Shift, Strong Emotional Reaction, or Impatience). It then inserts a placeholder tag \texttt{<interruption/>} at the precise logical point within the Assistant's text.
    \item \textbf{Execution Phase:} Conditioned on the chosen motivation and the context prior to the interruption point, the User Agent generates the specific interruption utterance.
\end{enumerate}

\paragraph{Backchannel Injection.}
Backchannels are injected probabilistically ($p=0.5$) during post-processing.
\begin{itemize}
    \item \textbf{User Backchannels:} The User Agent reviews the Assistant's response to insert feedback signals (uh-huh, gotcha) wrapped in \texttt{<user\_backchannel>} tags.
    \item \textbf{AI Backchannels:} Similarly, the system generates backchannels for the User's speech to simulate active listening by the Assistant.
\end{itemize}

\subsection{Quality Control}
We employ a rigorous filtering process. A \textbf{Reviewer Agent} scores the final dialogue on a scale of 1-5 across three dimensions: Persona Consistency, Quality of Interruption Event, and Naturalness of Backchannels. Dialogues with low logical consistency or failed event executions are discarded.
\subsection{Prompt Templates}
\label{sec:appendix_prompts}

We provide the core system prompts used in our pipeline below.

    \begin{promptbox}{Prompt 1: User Role-Play Instruction}
    \textbf{System Instruction:}
    You are a person in a real-time voice conversation. In the conversation history, your lines are marked with "speaker": "You". You are talking to the person marked "speaker": "Other".
    
    \textbf{You are NOT an AI assistant.} Your task is to speak naturally based on your persona. React authentically, don't try to be helpful.
    
    \textbf{Persona Details:}
    \begin{itemize}[leftmargin=*, nosep, label={-}] 
        \item \textbf{Persona:} \{persona\}
        \item \textbf{Communication Style:} \{style\}
    \end{itemize}
    
    \textbf{Conversation Rules:}
    - Speak, don't write: Use filler words (e.g., "um", "uh", "like"), hesitations, and natural phrasing.
    - Stay in character: Let your persona guide your responses.
    
    \textbf{Task:}
    Now, it's your turn. Generate your next response as "You".
    \end{promptbox}

    \begin{promptbox}{Prompt 2: Interruption Planning (Motivation \& Placement)}
    \textbf{Context:} The assistant is currently saying: “\{context\}”
    
    \textbf{Task: Plan and Place the Interruption}
    Your goal is to find the perfect moment to interrupt, driven by your persona.
    
    \textbf{Action 1: Plan the Interruption's Motivation.}
    First, think about \textit{why} your persona would interrupt here. Choose a motivation that fits your character:
    \begin{itemize}
        \item Correction: The assistant's response contains a point that may not be entirely accurate, and you want to clarify or refine it.
        \item Deeper Inquiry: You need to ask for clarification on a key point before they move on.
        \item Topic Shift: What they said reminds you of something else, and you want to change the subject.
        \item Strong Emotional Reaction: You are surprised, excited, or disagree strongly and can't hold it in.
        \item Impatience: You want to cut to the chase or stop a lengthy explanation.
    \end{itemize}
    
    \textbf{Action 2: Place the Interruption Marker.}
    Based on your chosen motivation, find the most natural point in the assistant's speech to jump in. Insert ONLY the empty tag pair \texttt{<interruption></interruption>} at that precise spot.
    \end{promptbox}

    \begin{promptbox}{Prompt 3: Interruption Utterance Generation}
    \textbf{Context:} You just decided to interrupt the assistant while they were saying: “\{context\}”
    Your motivation for interrupting is: \textbf{\{motivation\}}
    
    \textbf{Task: Deliver the Interruption}
    Now, say the words you would use to interrupt. Your utterance must sound spontaneous and directly reflect your motivation.
    
    [Detailed examples for motivations provided to the model: Correction, Deeper Inquiry, Topic Shift, Strong Emotional Reaction, Impatience]
    
    \textbf{Output:}
    Generate ONLY the interrupting phrase itself.
    \end{promptbox}

    \begin{promptbox}{Prompt 4: User Backchannel Generation}
    \textbf{Task:} As the assistant is speaking, you want to show you're listening. The assistant's last utterance was: “\{context\}”.
    \begin{itemize}
        \item \textbf{Action 1:} Think of a short, spoken backchannel phrase (e.g., "uh-huh", "gotcha", "right", "mhm").
        \item \textbf{Action 2:} Find the most natural point in the assistant's speech to insert this backchannel, wrapped in \texttt{<user\_backchannel>} tags.
    \end{itemize}
    \end{promptbox}

    \begin{promptbox}{Prompt 5: Final Dialogue Quality Review}
    \textbf{Role:} You are a meticulous evaluator of simulated spoken dialogues.
    
    \textbf{Scoring Criteria:}
    \begin{enumerate}
        \item \textbf{Persona Consistency \& Depth (1-5):} Does the user's speech effectively embody the assigned persona (\{persona\}) and style (\{style\})?
        \item \textbf{Quality of Interruption Event (1-5):} Is the interruption timed perfectly and motivated by the persona? Does it feel natural or forced?
        \item \textbf{Naturalness of Backchannels (1-5):} Are backchannels subtle and placed to improve flow, or are they distracting/robotic?
    \end{enumerate}
    
    \textbf{Output:} Provide scores and a detailed justification explaining your reasoning.
    \end{promptbox}

\end{document}